%% file: main.tex
\documentclass{article}

\usepackage{iclr2026_conference,times}

\input{math_commands.tex}

\usepackage{hyperref}
\usepackage{lineno}

\usepackage{url}
\usepackage{float}
\usepackage{multirow}
\usepackage{booktabs}

\usepackage[utf8]{inputenc} 
\usepackage[T1]{fontenc}    
\usepackage{hyperref}       
\usepackage{url}            
\usepackage{booktabs}       
\usepackage{amsfonts}       
\usepackage{array}
\usepackage[export]{adjustbox}
\usepackage{multirow}
\usepackage{makecell}

\usepackage{graphicx}       
\usepackage{nicefrac}       
\usepackage{microtype}      
\usepackage{xcolor}         
\usepackage{amsfonts}       
\usepackage{amsmath}        
\usepackage{lipsum}
\usepackage{caption}
\usepackage{subcaption}

\newcommand{\reb}[1]{\textcolor{black}{#1}}
\newcommand{\dataset}{GIQ}

\title{\dataset: Benchmarking 3D Geometric Reasoning of Vision Foundation Models with Simulated and Real Polyhedra}

\author{
\centerline{\textbf{Mateusz Michalkiewicz}$^1$,
\textbf{Anekha Sokhal}$^1$,
\textbf{Tadeusz Michalkiewicz}$^2$,
\textbf{Piotr Pawlikowski}$^2$,}\\
\centerline{\textbf{Mahsa Baktashmotlagh}$^3$,
\textbf{Varun Jampani}$^4$,
\textbf{Guha Balakrishnan}$^1$}\\[0.2cm]
\centerline{$^1$Rice University} \\
\centerline{$^2$Independent Researcher} \\
\centerline{$^3$The University of Queensland} \\
\centerline{$^4$Arcade AI}
}

\iclrfinalcopy 
\begin{document}

\maketitle

\input{0-abstract}

\input{1-introduction}

\input{2-related_work}

\input{3-dataset}
\input{4-experiments}

\input{5-conclusion}



\bibliography{main}
\bibliographystyle{iclr2026_conference}

\appendix
\clearpage

\end{document}

%% file: math_commands.tex
\usepackage{amsmath,amsfonts,bm}

\def\eqref#1{equation~\ref{#1}}

\def\1{\bm{1}}

\DeclareMathAlphabet{\mathsfit}{\encodingdefault}{\sfdefault}{m}{sl}
\SetMathAlphabet{\mathsfit}{bold}{\encodingdefault}{\sfdefault}{bx}{n}



%% file: 0-abstract.tex
\begin{abstract}
Modern monocular 3D reconstruction methods and vision-language models (VLMs) demonstrate impressive results on standard benchmarks, yet recent works cast doubt on their true understanding of geometric properties. We introduce \dataset{}, a comprehensive benchmark specifically designed to evaluate the geometric reasoning capabilities of vision and vision-language foundation models. \dataset{} comprises synthetic and real-world images and corresponding 3D meshes of diverse polyhedra covering varying levels of complexity and symmetry, from Platonic, Archimedean, Johnson, and Catalan solids to stellations and compound shapes. Through systematic experiments involving monocular 3D reconstruction, 3D symmetry detection, mental rotation tests, and zero-shot shape classification tasks, we reveal significant shortcomings in current models. State-of-the-art reconstruction algorithms trained on extensive 3D datasets struggle to reconstruct even basic geometric Platonic solids accurately. Next, although foundation models may be shown via linear \reb{and non-linear} probing to capture specific 3D symmetry elements, they falter significantly in tasks requiring detailed geometric differentiation, such as mental rotation.
Moreover, advanced vision-language assistants such as ChatGPT, Gemini and Claud exhibit remarkably low accuracy in interpreting basic shape properties such as face geometry, convexity, and compound structures of complex polyhedra. \dataset{} is publicly available at \textcolor{blue}{\url{toomanymatts.github.io/giq-benchmark/}}, providing a structured platform to benchmark critical gaps in geometric intelligence and facilitate future progress in robust, geometry-aware representation learning.
\end{abstract}

%% file: 1-introduction.tex
\section{Introduction}
Computer vision models trained on massive image datasets now achieve state-of-the-art performance on a range of visual reasoning tasks. However, unlike humans who naturally reason about the world in terms of 3D structure, these models learn to exploit arbitrary (often 3D-unaware) patterns discovered from their training datasets. This property can, in turn, lead to poor generalization performance on out-of-distribution data. For example, vision-language models (VLMs) are known to struggle with questions related to depth ordering ~\citep{tong2024cambrian}, and as shown in our experiments, monocular reconstruction algorithms struggle to reconstruct shapes outside of their training distributions. For this reason, there is growing interest in understanding the 3D ``awareness'' of popular foundation vision models, such as by ``linearly probing'' ~\citep{alain2017understanding} their intermediate features to solve basic 3D reasoning tasks such as depth estimation and scene registration ~\citep{el2024probing,man2024lexicon3d}. 

These few existing 3D analysis studies rely on large datasets of synthetic or in-the-wild objects such as OmniObject3D~\citep{wu2023omniobject3d}, NAVI~\citep{jampani2023navi}, and Google Scanned Objects (GSO)~\citep{downs2022google}. While these datasets enable large-scale analyses, they are not conducive to careful evaluation of visual reasoning with respect to core 3D object geometry properties such as symmetry, convexity, and complexity. For example, is CLIP~\citep{radford2021learning} better than DINO~\citep{oquab2023dinov2} at identifying center-point symmetries, and if so, to what degree? And how do reconstructions of SF3D~\citep{boss2024sf3d} and OpenLRM~\citep{hong2023lrm} compare as shape complexity increases? Quantitative answers to such questions can provide crucial information to benchmark and ultimately improve 3D geometric reasoning capabilities of these algorithms. 

\begin{figure}[t!]
  \centering
  \includegraphics[width=\textwidth, trim=0 65 0 0, clip]{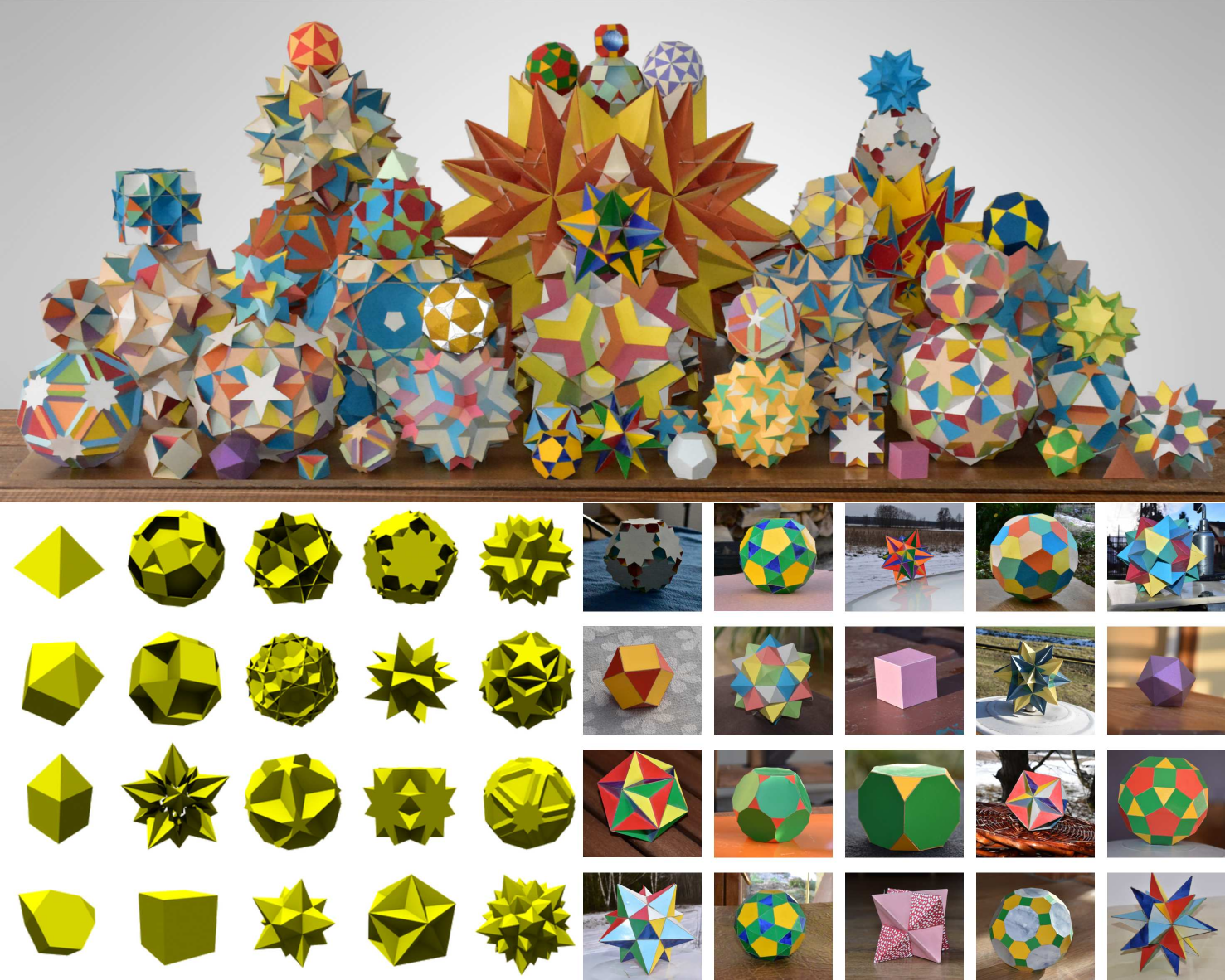}
  \caption{\textbf{Samples of synthetic and real 3D solids from our \dataset{} dataset.} 
  A subset of the 224 real polyhedra included in our dataset, illustrating their variety in complexity, class, and colors. (bottom left) Simulated solids from Mitsuba Physically Based Renderer. (bottom right) Real polyhedra constructed from paper, placed in different realistic backgrounds.}
  \label{fig:ds_overview}
\end{figure}

In this study, we introduce \dataset{} (for \emph{Geometric IQ} Test), a first-of-its-kind dataset consisting of simulated and physical polyhedra to help answer such questions (see Fig.~\ref{fig:ds_overview}). Polyhedra have fascinated mathematicians for centuries due to their fundamental nature in geometry, inspiring classical works such as Plato's association of regular solids with classical elements, Kepler's exploration of polyhedral properties in his \textit{Harmonices Mundi} \citep{kepler1969harmonices}, and Euler's groundbreaking analysis of polyhedral topology \citep{euler1758elementa}. 
Polyhedra also appear in many practical scientific applications, from crystalline minerals (e.g., cubic iron pyrite) and molecular structures in nature (e.g., the truncated icosahedral ``buckyball'') to fundamental primitives in graphics algorithms.

Unlike arbitrary shapes, the well-defined categories of polyhedra (e.g., Platonic, Archimedean, Johnson solids) along with their precise symmetry groups (e.g., tetrahedral, octahedral, icosahedral) provide unambiguous ground truth for evaluation. Furthermore, polyhedra exhibit a rich hierarchy of geometric complexity -- from simple Archimedean solids such as the tetrahedron with four faces to 
the highly complex great dirhombicosidodecahedron, nicknamed \textit{Miller's Monster}~\citep{verheyen1989complete}, a nonconvex uniform polyhedron with 124 faces (40 triangles, 60 squares, and 24 pentagrams) 
-- enabling systematic investigation of how shape complexity affects perception. 
The perceptual salience of these patterns makes polyhedra an excellent testbed for vision models. 

To this end, we constructed \dataset{} with images of 224 unique polyhedra captured from multiple viewpoints. We constructed synthetic polyhedra using the Mitsuba Physically Based Renderer \citep{nimier2019mitsuba} (see Fig.~\ref{fig:ds_overview} bottom left), and constructed intricate physical polyhedra models from paper and placed them in various indoor and outdoor real-world environments (see Fig.~\ref{fig:ds_overview} bottom right). This is the first dataset of its kind with such a quantity and range of synthetic and real polyhedra. 

We used \dataset{} to perform systematic evaluations on a range of state-of-the-art vision-language models and monocular 3D reconstruction methods to assess their capability to recognize symmetry, reconstruct complex geometries from a single view, and accurately reason about shape equivalence across diverse viewpoints and real-world conditions. We find that even for the simplest Platonic solids, single‑image reconstruction is unreliable. Across state‑of‑the‑art monocular 3D reconstruction methods trained on millions of diverse 3D assets, we observe frequent failures to recover even the properties of a cube: axis‑aligned faces, right angles, uniform edge lengths, and underlying 3D symmetries (e.g., 4‑fold rotational invariances), leading to asymmetric artifacts and mismatched global structure across views. Furthermore, as geometric complexity increases to nonconvex or compound forms, reconstruction quality degrades further, exhibiting fractured surfaces and topological inconsistencies.  Interestingly, while we found that a lightweight \reb{non-linear} probe on foundation image encoders (e.g., \reb{SigLip}, DINOv2) can reliably classify symmetry cues, 
we also observed that these encoders and frontier vision–language assistants (ChatGPT, Gemini, Claude) struggle in classifying geometrically similar shapes. 
Finally, in zero‑shot shape classification, vision-language assistants achieve below 20\% accuracy on simple Catalan and Johnson solids and on non‑convex shapes, frequently misidentifying face geometry, confusing convexity, or conflating compounds.

Results show that \dataset{} provides a targeted benchmark to diagnose fundamental geometric intelligence gaps in vision systems, laying the groundwork for future principled improvements in spatial perception and 3D-aware visual reasoning.

%% file: 2-related_work.tex
\section{Related Work}

Foundation vision models, such as CLIP~\citep{radford2021learning}, DINO~\citep{oquab2023dinov2}, and Llama~\citep{touvron2023llama}, have achieved widespread popularity due to their remarkable performance across diverse vision tasks. Given their widespread adoption, recent studies increasingly investigate these models' robustness and generalization properties, including sensitivity to adversarial perturbations~\citep{schlarmann2023adversarial}, reliance on dataset biases~\citep{nguyen2022quality}, capacity for compositional reasoning~\citep{doveh2023dense,lewis2022does}, and resilience to visual anomalies~\citep{zhu2024llms}. Among these efforts, there is a growing emphasis on evaluating how effectively these models encode and reason about 3D structures~\citep{el2024probing,man2024lexicon3d}, spatial relations~\citep{ramakrishnan2024does,tang2024sparkle,yamada2023evaluating,yang2023set}, and spatial cognition more broadly.

A particularly critical aspect of spatial cognition studied extensively in cognitive science is the ability to recognize objects under rotation. The Mental Rotation Test (MRT), first proposed by ~\cite{shepard1971mental} and later standardized by ~\cite{vandenberg1978mental}, assesses spatial reasoning by requiring subjects to determine whether two rotated 3D objects are identical. 
Recent work by ~\cite{ramakrishnan2024does} examined mental rotation using synthetic stimuli consisting of Lego-like blocks. Building upon this, our study expands the MRT evaluation framework by incorporating polyhedral shapes of varying complexity and geometric properties, evaluated across both synthetic renderings and real-world images.

Another fundamental aspect of spatial intelligence extensively studied in cognitive science is symmetry perception. Humans exhibit a well-documented sensitivity and preference for symmetric patterns~\citep{wagemans1997characteristics}. In computer vision, symmetry detection has been explored both in two-dimensional contexts~\citep{tsogkas2012learning,funk2017beyond} and, more recently, extended into three-dimensional settings~\citep{gao2020prs}.
Despite substantial advancements, existing large-scale 3D datasets like Objaverse~\citep{deitke2023objaversee} and its successor Objaverse XL~\citep{deitke2023objaverse} typically lack detailed annotations necessary to evaluate fine-grained geometric reasoning, such as recognition of specific symmetry groups or subtle shape complexities. Addressing this gap, we introduce a targeted polyhedron-based benchmark, providing clearly defined ground truths on symmetry, complexity, and geometric properties, thereby enabling precise measurement of geometric understanding in foundation vision models.

The choice of polyhedra as our evaluation focus draws on a rich heritage of mathematical and scientific exploration. Historically, scholars such as Da Vinci, Descartes, Euler, and Gauss extensively explored these geometric forms, recognizing their structural elegance and conceptual clarity. In modern mathematics, polyhedra continue to be studied rigorously by scholars ~\citep{coxeter2012fifty,coxeter1954uniform,klein2003lectures,stewart1980adventures}. Building on this historical foundation, our dataset systematically leverages polyhedral geometry to rigorously probe and enhance the geometric intelligence of modern vision systems.

%% file: 3-dataset.tex
\section{Dataset}
We describe the composition of \dataset{} in this section. We first present geometrical concept definitions in Sec.~\ref{dataset:prelim}, followed by a description of all polyhedral shapes we considered in Sec.~\ref{dataset:shapes}. We then describe the synthetic and real instantiations of these shapes in \dataset{} in Sec.~\ref{dataset:synthetic} and Sec.~\ref{dataset:wild}.

\subsection{Geometry Preliminaries}
\label{dataset:prelim}
We first briefly define key geometric concepts related to polyhedra which will underpin the detailed discussion of polyhedral classes presented in the following subsections. Following Coxeter's definition, a polyhedron is a finite set of polygons arranged so that every side of each polygon belongs to exactly one other polygon, with no subset of polygons having this property (i.e. the entire collection is connected)~\citep{coxeter1954uniform}. A \emph{regular polygon} is equilateral (all sides equal) and equiangular (all internal angles equal). A polyhedron with faces that are all congruent regular polygons (i.e., identical in size and shape) 
is termed a \textbf{regular polyhedron}. A polyhedron is \textbf{convex} if any line segment joining two points inside or on its surface remains completely within or on its surface. Otherwise, it is considered \textbf{non-convex} (or concave). A polyhedron is \textbf{vertex-transitive} if any vertex can be mapped onto any other vertex by a symmetry operation (rotation, reflection, or translation).  Similarly, a polyhedron is \textbf{edge-transitive}/\textbf{face-transitive} if all edges/faces can be mapped onto each other by symmetry operations. A \textbf{uniform polyhedron} has regular polygonal faces and is vertex-transitive, meaning that the arrangement of its faces around each vertex is identical. Lastly, 
\textbf{dual polyhedra} are pairs of polyhedra where vertices and faces are interchanged: vertices of one polyhedron correspond exactly to faces of the other, and vice versa. 

\input{table_stats}

\subsection{3D Shapes}
\label{dataset:shapes}
We considered a comprehensive collection of polyhedral classes systematically varying in geometric complexity, symmetry properties, and topological regularity. We began with the \textit{Platonic solids}, comprising exactly five regular convex polyhedra: tetrahedron, cube, octahedron, dodecahedron, and icosahedron. Relaxing the condition that all faces must be identical but retaining vertex-transitivity and edge-transitivity yields the broader set of \textit{Archimedean solids}, consisting of 13 convex polyhedra whose faces are regular polygons of more than one type, arranged symmetrically around each vertex. Taking the dual polyhedra of the Archimedean solids leads to the 13 \textit{Catalan solids}~\citep{catalan1865memoire}. 
Each Catalan solid has congruent faces but allows variation in vertex configurations. Removing the requirement for vertex and face transitivity entirely produces the family of \textit{Johnson solids}, comprising 92 strictly convex polyhedra with regular polygonal faces but lacking  vertex uniformity~\citep{Johnson1966,zalgaller1969convex}. To further enrich the diversity of shapes in our dataset, we incorporated \textit{stellations}, geometric constructions formed by extending the faces or edges of polyhedra until they intersect again, creating more complex, nonconvex forms. We included selected stellations of the octahedron, dodecahedron, icosahedron, cuboctahedron, and icosidodecahedron~\citep{coxeter2012fifty}.  Additionally, we included various \textit{compound polyhedra}: structures formed by the symmetric combination of multiple polyhedra (e.g., compound of cube and octahedron, and compound of dodecahedron and icosahedron). Both stellations and compound polyhedra significantly enhance structural complexity, introducing intricate symmetry properties and challenging visual configurations. Finally, we considered the complete set of polyhedra described by \cite{Wenninger1971}, encompassing 119 shapes in total. Beyond Platonic, Archimedean, and stellations, this set includes the \textit{Kepler-Poinsot solids}, and a collection of nonconvex polyhedra exhibiting a wide array of geometric and topological complexities.

In total, the \dataset{} dataset consists of 224 carefully curated 3D shapes, from symmetric and simple regular forms to intricate, nonconvex, and irregular structures, providing a robust basis for exploring geometric perception and spatial reasoning in neural models. We summarize counts of specific shapes in the dataset in Figure~\ref{fig:polyhedra_summary} with additional representative samples and listed key geometric features for each group provided in the appendix.

\subsection{Synthetic Renderings}
\label{dataset:synthetic}
To systematically evaluate spatial reasoning under controlled conditions, we rendered each of the 3D shapes described above using the Mitsuba physically-based renderer~\cite{nimier2019mitsuba} from 20 randomly sampled viewpoints. We distributed these viewpoints uniformly over the viewing hemisphere, ensuring diverse perspective coverage. We used a perspective camera with a resolution of $256 \times 256$ pixels, a near clipping plane of $10^{-3}$, and a far clipping plane of $10^8$. For each view, the object is rendered using diffuse shading. To simulate realistic lighting while preserving shape detail, we used a two-sided diffuse BRDF with a high-reflectance yellowish surface. Each rendering uses 1024 low-discrepancy samples per pixel to ensure smooth convergence.
Instead of global illumination, which tended to reduce contrast between adjacent faces and obscure geometric boundaries, we adopted a direct integrator with four emitter samples and no BSDF sampling. This setup emphasizes sharp direct shading cues and produces face-to-face contrast more consistent with wild images, where distinguishing neighboring facets is typically easier.

\subsection{Wild Images}
\label{dataset:wild}
To complement our synthetic renderings with real-world variabilities, we placed physical paper models of the polyhedra---constructed by Piotr Pawlikowski---in natural settings.
Our wild image dataset includes all Platonic, Archimedean, Catalan, and non-prismatic uniform polyhedra, as well as the Kepler–Poinsot solids and a representative subset of 48 Johnson solids. These paper models were manually assembled following geometric blueprints.

We photographed each shape using a Nikon D3500 DSLR camera at a native resolution of 6000$\times$4000 pixels under two broad conditions.  First, we captured approximately 20 indoor images per shape under controlled lighting. Second, we collected 20 outdoor images per shape under varying environmental conditions, including sunny weather, overcast skies, and snowy winter backgrounds. These outdoor images introduce rich natural variability in illumination, background texture, and color. We randomly selected viewpoints to ensure full 360-degree coverage, while maintaining sufficient visibility of shape geometry. We provide a visual overview of our wild image dataset in Figure~\ref{fig:ds_overview}\reb{; additional examples are provided in the appendix}.

%% file: table_stats.tex
\begin{figure}[t]
\centering
\newcommand{\imgwidth}{0.33\linewidth}
\newcommand{\imgheight}{1.2cm}
\newcommand{\polyimg}[1]{%
  \makebox[\imgwidth][c]{\includegraphics[width=\imgwidth,height=\imgheight]{#1}}%
}

\begin{tabular}{|
>{\centering\arraybackslash}m{0.15\linewidth}|
>{\centering\arraybackslash}m{0.28\linewidth}|
>{\centering\arraybackslash}m{0.15\linewidth}|
>{\centering\arraybackslash}m{0.28\linewidth}|}
\hline
\textbf{Group [\#]} & \textbf{Examples} & \textbf{Group [\#]} & \textbf{Examples} \\ \hline

\small Platonic [5] &
{%
\raisebox{-0.15cm}{%
\polyimg{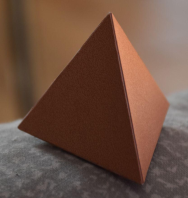}%
\polyimg{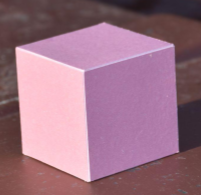}%
\polyimg{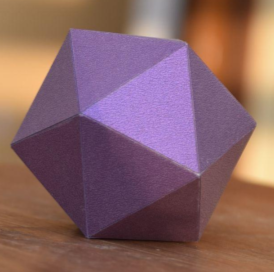}%
}%
} &

\small Archimedean~[13] &
{%
\raisebox{-0.15cm}{%
\polyimg{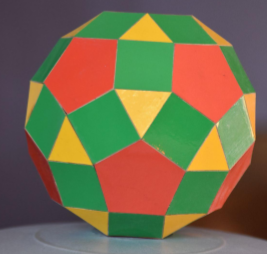}%
\polyimg{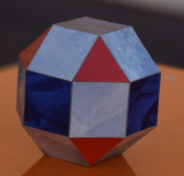}%
\polyimg{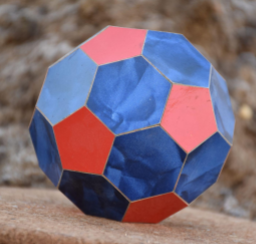}%
}} \\ \hline 

\small Catalan~[13] &
{\raisebox{-0.15cm}{%
\polyimg{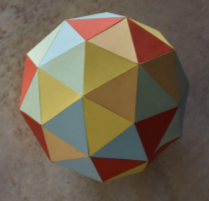}%
\polyimg{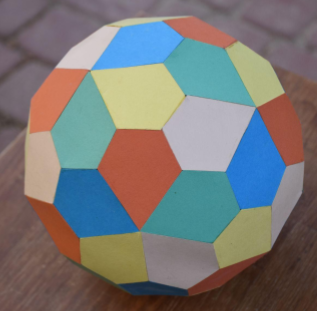}%
\polyimg{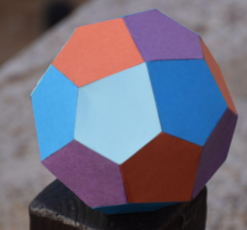}%
}} &

\small Johnson~[92] &
{\raisebox{-0.15cm}{%
\polyimg{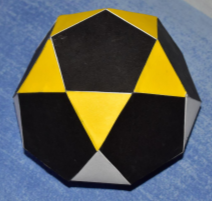}%
\polyimg{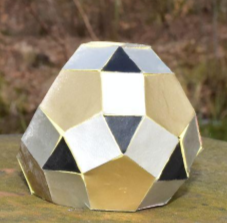}%
\polyimg{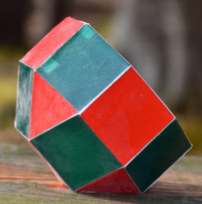}%
}} \\ \hline
\small Stellations~[48] &
{%
\raisebox{-0.15cm}{%
\polyimg{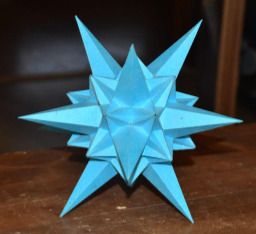}%
\polyimg{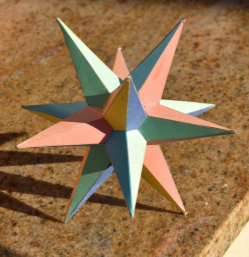}%
\polyimg{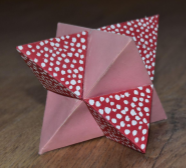}%
}} &

\small Kepler-Poinsot~[4] &
{%
\raisebox{-0.15cm}{%
\polyimg{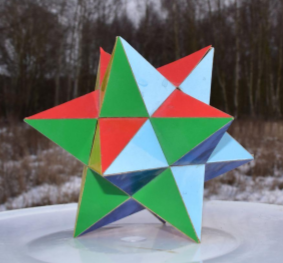}%
\polyimg{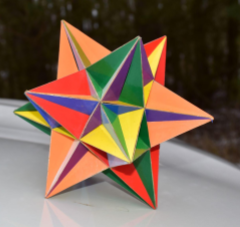}%
\polyimg{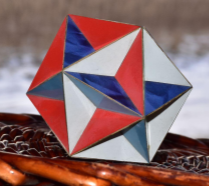}%
}} \\ \hline
\small Compounds~[10] &
{%
\raisebox{-0.15cm}{%
\polyimg{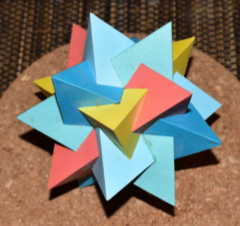}%
\polyimg{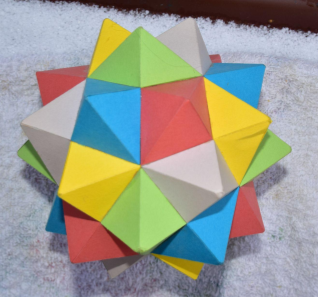}%
\polyimg{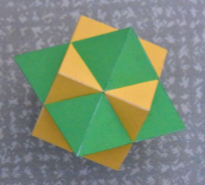}%
}} &

\small Uniform non-convex~[53] &
{%
\raisebox{-0.15cm}{%
\polyimg{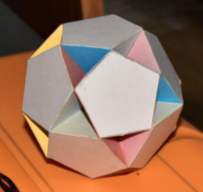}%
\polyimg{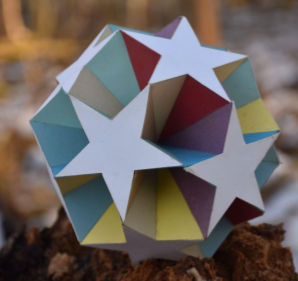}%
\polyimg{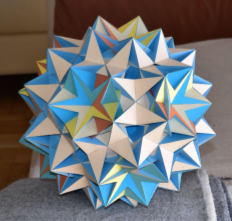}%
}} \\ \hline

\end{tabular}

\caption{\textbf{Summary of polyhedral groups in GIQ, highlighting group names, counts of distinct 3D shapes (in parentheses), and representative examples.} Platonic, Archimedean, and Catalan solids are convex, while Kepler-Poinsot polyhedra and compounds represent special cases of stellations; \reb{consequently, the sum of group counts (238) exceeds the 224 unique shapes in the dataset.} The categorization presented here is arbitrary: polyhedra possess numerous properties allowing various groupings; we selected this set as a representative example.}
\label{fig:polyhedra_summary}
\end{figure}

%% file: 4-experiments.tex
\section{Experiments}
We used the \dataset{} dataset to conducted a series of comprehensive experiments to evaluate the geometric reasoning capabilities of contemporary foundation vision models. We focused on four types of experiments: \textit{Monocular 3D Reconstruction}, \textit{3D Symmetry Detection}, \textit{Mental Rotation Tests}, and \textit{Zero-Shot Polyhedron Classification}. These experiments are designed to probe different dimensions of geometric intelligence: explicit 3D reconstruction of shapes from single images, implicit embedding-based detection of 3D symmetries and subtle geometric distinctions, and high-level semantic classification of frontier vision-language models. \reb{We note that while ``wild'' images retained their complex backgrounds and varying lighting conditions for Symmetry, Mental Rotation, and Classification tasks to test environmental robustness, they were preprocessed (background removal and centering) exclusively for the Monocular 3D Reconstruction task to align with the input assumptions of baseline methods. Detailed quantitative results, separated by synthetic and wild domains, are provided in the appendix}. Collectively, these evaluations highlight both the current strengths and shortcomings of state-of-the-art models to represent, recognize, and reason about complex geometric structures.

\subsection{Monocular 3D Reconstruction}
We first evaluated three recent state-of-the-art monocular 3D reconstruction methods: \textbf{Shap-E}~\citep{jun2023shap}, \textbf{Stable Fast 3D}~\citep{boss2025sf3d}, and \textbf{OpenLRM}~\citep{openlrm}.
Shap-E is a diffusion-based model trained on millions of 3D assets. It encodes shapes with a Transformer and generates textured meshes via implicit neural representations.
Stable Fast 3D is built on TripoSR~\citep{tochilkin2024triposr}, encodes images with DINOv2~\citep{oquab2023dinov2} and outputs triplane-based representations. It is trained on a curated subset of Objaverse~\citep{deitke2023objaversee}.
OpenLRM, based on LRM~\citep{honglrm}, predicts neural radiance fields from single images using a transformer, and is trained end-to-end on Objaverse and MVImgNet~\citep{yu2023mvimgnet}. \reb{We evaluated these pre-trained models zero-shot, using \dataset{} as an out-of-distribution test set.}

The experimental task involves reconstructing the complete 3D geometry of polyhedral shapes from a single image. We selected representative examples across three categories: cube (Platonic solid), great dodecahedron (Kepler-Poinsot solid), and small cubicuboctahedron (uniform nonconvex solid).
To provide a fair comparison, we preprocessed wild images via center cropping and background removal. Table ~\ref{tab:m3dr-threeviews} presents results of this evaluation. Despite extensive training on millions of diverse shapes or inclusion of similar solids in the training dataset, all three models exhibited significant reconstruction failures for most tested cases, particularly on wild images and shapes beyond the simplest forms. Shap-E successfully reconstructed the synthetic cube but failed dramatically on the wild cube and more complex shapes. Stable Fast 3D accurately captured front-facing geometries for several shapes but consistently failed to coherently reconstruct side and rear views. Similarly, OpenLRM produced plausible reconstructions for simple synthetic shapes but exhibited considerable inaccuracies across viewpoints and struggled substantially with complex and wild imagery. Additional qualitative and quantitative reconstruction results are provided in the appendix.

\input{table_m3dr}

\subsection{3D Symmetry Detection}
We evaluated the capability of various image encoders to detect specific 3D symmetry elements from single-image inputs. The task involved predicting the presence of three distinct symmetry elements in objects depicted by images: \reb{central point reflection (invariance under inversion through a central point), and $n$-fold rotational symmetries, defined as invariance under rotation by $360^{\circ}/n$. Specifically, we target 4-fold ($90^{\circ}$) and 5-fold ($72^{\circ}$) symmetries.} . Unlike common evaluations focused on 2D planar symmetries or image rotations, our evaluation explicitly targets recognition of inherent symmetry elements defined with respect to the object's three-dimensional structure.

We examined a diverse set of encoders with variations in modalities, architectures, and supervision strategies. Specifically, we selected six supervised methods, three image-and-text transformer-based models (CLIP~\citep{radford2021learning}, DreamSim~\citep{fu2023dreamsim}, SigLip~\citep{zhai2023sigmoid}), and three image CNN-based ones (DeiT III~\citep{touvron2022deit}, SAM~\citep{foret2020sharpness, chen2021vision}, and ConvNext~\citep{liu2022convnet}). Additionally, we considered three self-supervised transformer-based methods trained solely on images: DINO~\citep{caron2021emerging}, DINOv2~\citep{oquab2023dinov2}, and Masked AutoEncoder (MAE)~\citep{he2022masked}. \reb{Finally, to assess geometry-native representations, we included three multi-view pretrained networks: VGGT~\citep{wang2025vggt}, DUSt3R~\citep{wang2024dust3r}, and MASt3R~\citep{leroy2024grounding}.}

We applied a linear probe~\citep{alain2017understanding}, feeding each featurizer’s embeddings into a single linear layer to classify symmetry elements. \reb{To assess if geometric information was encoded in non-linearly separable manifolds, we also evaluated non-linear probes; however, these yielded comparable average performance to linear probes (see appendix).}
To address class imbalance, we employed a weighted binary cross-entropy loss. For each symmetry class \( c \), we compute the positive-class weight as $w_c = \frac{N - n_c}{n_c}$, where \( n_c \) is the number of positive samples for class \( c \) and \( N \) is the total number of samples. We then define the per-example loss for logits \( z_{i,c} \) and targets \( y_{i,c}\in\{0,1\} \) as:
\[
  \mathcal{L} = -\frac{1}{N}\sum_{i=1}^N\sum_{c=1}^C 
  \left[ w_c\,y_{i,c}\log\sigma(z_{i,c}) 
  + (1 - y_{i,c})\log(1 - \sigma(z_{i,c})) \right],
\]
where \( \sigma(z_{i,c}) \) denotes the predicted probability for class \( c \) on example \( i \). We used balanced accuracy, computed as
$ 0.5 \cdot \frac{\text{TP}}{P} + 0.5 \cdot \frac{\text{TN}}{N} $,
where TP and TN represent true positives and true negatives respectively (with \( P \) and \( N \) being the number of positive and negative samples), as our primary evaluation metric.

We trained models on embeddings from synthetic images and subsequently evaluated them on embeddings extracted from both synthetic and real-world (wild) images. We excluded Johnson solids due to ambiguities arising from certain viewing angles. \reb{To ensure robust performance estimates and mitigate potential biases from specific dataset splits, we employed a 5-fold cross-validation strategy. We partitioned the unique polyhedral shapes into five disjoint folds, ensuring that in every iteration, the test set contained only shapes unseen during training. Detailed information about dataset composition is provided in the appendix.}

\reb{We present the averaged balanced accuracy across all five} folds in Figure~\ref{fig:mrt_results} (left). Notably, DINOv2 consistently delivered \reb{strong} performance across symmetry categories and particularly excelled in the recognition of 4-fold rotational symmetry, achieving up to 93\% accuracy on wild images despite being trained only on synthetic data.

\subsection{Mental Rotation Test}
Next, we evaluated foundation image encoders' spatial reasoning capabilities through a Mental Rotation Test (MRT)~\citep{shepard1971mental}. This task involves determining whether two images-one synthetic rendering and one real-world (wild) photograph-depict the same polyhedral object, differing only by rotation. Such a scenario is relevant to real-world applications, for instance, a robot trained on CAD models needing to locate a physical object with matching geometric properties in an unstructured environment. 

We first used an 80\%-20\% train-test split with synthetic image pairs alone to establish baseline performance. Under these simplified conditions, model accuracies ranged between 93\%-98\% (complete results available in appendix). Through this initial experiment, we identified that combining image embeddings using absolute difference ($|e_1 - e_2|$) followed by a non-linear probe yielded the highest accuracy. We further explored alternative embedding combination methods (concatenation and subtraction) 
\reb{and linear probing baselines with detailed results per featurizer reported in the appendix} 
However, since the trivial setting posed insufficient challenges for discerning nuanced differences, we subsequently introduced a more demanding \textit{hard} split. This split is specifically designed with visually and geometrically similar polyhedra pairs, rigorously testing the models' fine-grained shape differentiation abilities. Examples of these challenging shape pairs are shown in the appendix.

\begin{figure}
  \centering
  \includegraphics[width=\textwidth, trim=0cm 3.1cm 0cm 2.5cm, clip]{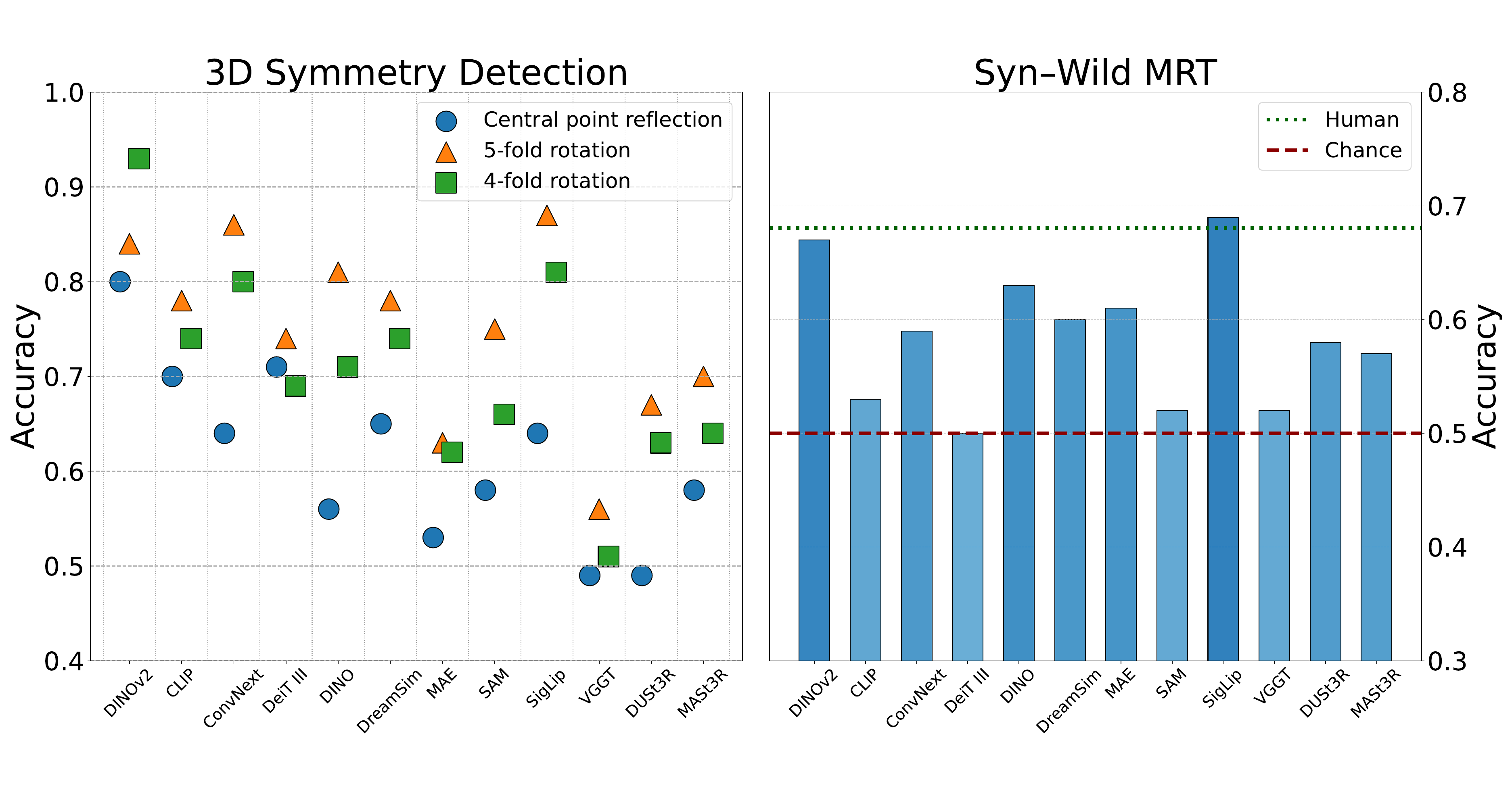}
\caption{Left: Balanced accuracy (\(0.5 \cdot \frac{\text{TP}}{P} + 0.5 \cdot \frac{\text{TN}}{N}\)) for linear probing of 3D symmetry detection using embeddings from different featurizers. The linear classifier is trained only on synthetic images (Syn), and evaluated on real-world (Wild) images for detecting three symmetry types: central point reflection, 5-fold rotation, and 4-fold rotation. Right: Mental Rotation Test accuracy using \reb{non-linear probes. Top models (e.g., SigLIP) match the human average ($\sim$69\%, green dotted line), though 68\% of human participants still outperformed the best model.}}
  \label{fig:mrt_results}
\end{figure}

Results for the \textit{hard} split are summarized in Figure~\ref{fig:mrt_results} (right). We trained on synthetic images and evaluated on both synthetic and synthetic-wild pairs, with syn-syn results provided in the appendix. All models exhibited a significant performance drop compared to the trivial split, particularly in the synthetic-wild test scenario, where average performance approached chance level. 
\reb{While non-linear probes enabled SigLIP and DINOv2 to achieve 69\% and 67\% accuracy respectively, the majority of models exhibited limited spatial reasoning capabilities, often struggling to surpass 60\%,} highlighting the considerable difficulty models face in reliably distinguishing subtle geometric differences.

\reb{To determine the extent to which these geometric cues are perceptible to human observers versus current vision models, we established a human baseline. We conducted a user study with 42 participants using a web interface to evaluate the exact image pairs from our test set. Participants answered 25 questions (5 from the ``easy'' split and 20 from the ``hard'' split) with a strict 30-second time limit per question. On the easy set, participants achieved a mean accuracy of 97.56\%, confirming task comprehension. On the hard set, human accuracy averaged 68.05\% (std dev: 0.11), with top performers scoring 90\% (18/20). Notably, while our best-performing model configuration (SigLIP with a non-linear probe) achieved parity with the average human ($\sim$69\%), 68\% of individual participants outperformed the best model.}

\subsection{Zero-Shot Polyhedron Classification by Frontier Models}
Finally, we conducted a zero-shot polyhedron classification task to evaluate the geometric reasoning capabilities of frontier vision-language models. Specifically, we assessed Claude 3.7 Sonnet~\citep{anthropic2025claude37}, Gemini 2.5 Pro~\citep{google2025gemini25pro}, ChatGPT o3, and ChatGPT o4-mini-high~\citep{openai2025o3} by querying each model with synthetic and real-world images from our dataset with the prompt: \textit{``What is the name of this polyhedron?"}. \reb{To ensure robust evaluation, we accounted for synonyms (e.g., Cube/Hexahedron, Stella Octangula/Stellated Octahedron) and manually verified that no models produced correct answers using synonyms outside our dictionary.} Classification accuracy across polyhedron categories is reported in Figure~\ref{fig:combined_cls} (a).

Model performance varied significantly across polyhedron categories. Gemini 2.5 Pro achieved perfect accuracy for Kepler-Poinsot solids, while ChatGPT o3 achieved a perfect score on Platonic solids. Conversely, all models struggled with Johnson solids, Catalan solids, uniform non-convex solids, and compound structures, indicating that even shapes with simple repetitive polygonal faces (e.g., Johnson or Catalan) remain difficult to classify, alongside the more complex non-convex cases. \reb{To investigate potential bottlenecks in prompting or input ambiguity, we extended our evaluation to include Chain-of-Thought (CoT) prompting and Multi-View (MV) inputs (providing three canonical views). CoT strategies yielded minimal gains, with models often hallucinating intermediate features. Similarly, MV inputs provided only marginal improvements, primarily for low-symmetry Johnson solids where single views can be ambiguous. For high-symmetry categories (Platonic, Archimedean, Compounds, Stellations), a single view is informationally complete; thus, the persistent failure indicates a lack of geometric reasoning rather than visual occlusion. Detailed results for these ablations are provided in the appendix.}

We qualitatively analyzed classification errors, revealing systematic patterns of geometric misinterpretation, and present input images with frontier-model reasoning in Figure~\ref{fig:combined_cls} (b); additional examples appear in the appendix. Models frequently (i) confused convex and concave structures—for example, \textbf{Claude 3.7 Sonnet} correctly described face types and coloring but still misclassified a concave shape as a cuboctahedron, an Archimedean solid that must be convex; (ii) conflated compounds with stellations or other multi-component assemblies—for instance, \textbf{ChatGPT o3} recognized a star-shaped compound but misidentified both the protrusions (triangular pyramids vs.\ octahedra) and the constituents (dodecahedron–icosahedron vs.\ a compound of octahedra), while \textbf{Gemini 2.5 Pro} similarly misclassified a color-coded cube–octahedron compound as a “stellated octahedron”; and (iii) hallucinated additional face types, as seen when \textbf{ChatGPT o4-mini-high} noted pentagonal faces but incorrectly added hexagons, ultimately blaming viewpoint ambiguity despite the full pattern being visible.

\reb{To determine if these limitations were specific to 2D-pretrained architectures, we extended our evaluation to geometry-native models. We tested LLaVA-3D \cite{zhu2024llava} (using image inputs), as well as ShapeLLM \cite{qi2024shapellm} and PointBind \& PointLLM \cite{guo2023point} (using ground truth point clouds as input). As shown in Figure~\ref{fig:combined_cls} (a), despite the availability of explicit 3D geometry, these models did not outperform the generalist frontier VLMs. Collectively, these limitations underscore a fundamental gap in current architectures, highlighting the necessity of enhanced mechanisms for explicit geometric representation to reliably differentiate complex structures.}

\input{table_0shot}

%% file: table_m3dr.tex
\newcommand{\imgwidth}{0.12\textwidth}

\begin{table}[h]
  \centering
  \setlength{\tabcolsep}{2pt}
  \begin{tabular}{|c|cc|cc|cc|}
    \hline
        \multirow{2}{*}{\textbf{Input}} & \multicolumn{2}{c|}{Shap-E } & \multicolumn{2}{c|}{Stable Fast 3D } & \multicolumn{2}{c|}{OpenLRM} \\
                   & Front & Side & Front & Side & Front & Side \\
    \hline
    \includegraphics[width=\imgwidth]{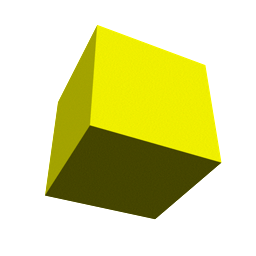} &
      \includegraphics[width=\imgwidth]{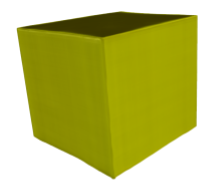} &
      \includegraphics[width=\imgwidth]{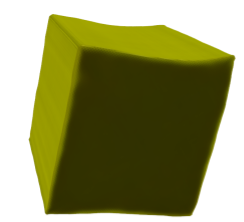} &
      \includegraphics[width=\imgwidth]{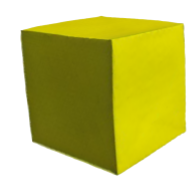} &
      \includegraphics[width=\imgwidth]{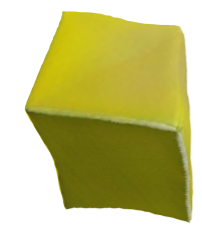} &
      \includegraphics[width=\imgwidth]{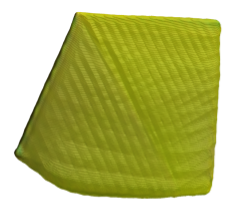} &
      \includegraphics[width=\imgwidth]{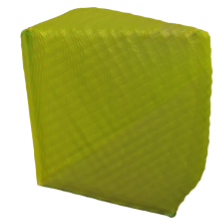} \\
    \hline
    \includegraphics[width=\imgwidth]{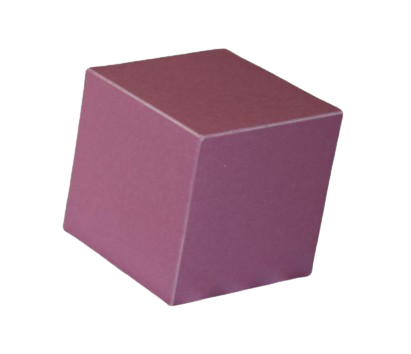} &
      \includegraphics[width=\imgwidth]{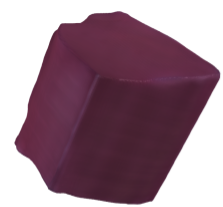} &
      \includegraphics[width=\imgwidth]{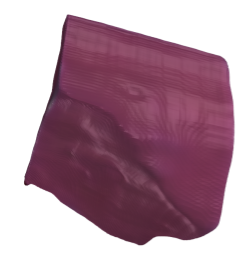} &
      \includegraphics[width=\imgwidth]{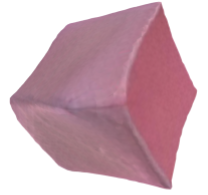} &
      \includegraphics[width=\imgwidth]{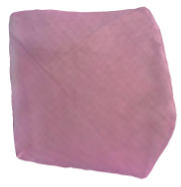} &
      \includegraphics[width=\imgwidth]{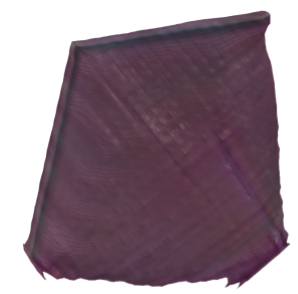} &
      \includegraphics[width=\imgwidth]{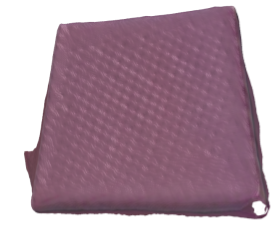} \\
    \hline
    \includegraphics[width=\imgwidth]{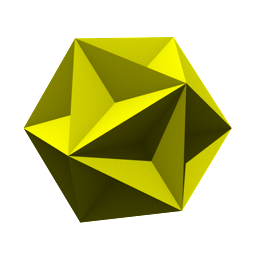} &
      \includegraphics[width=\imgwidth]{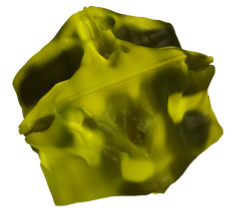} &
      \includegraphics[width=\imgwidth]{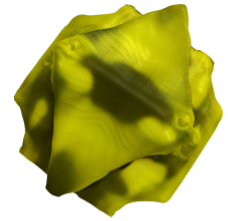} &
      \includegraphics[width=\imgwidth]{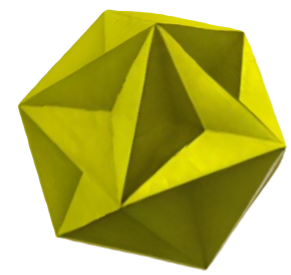} &
      \includegraphics[width=\imgwidth]{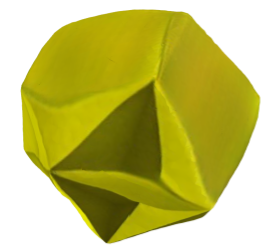} &
      \includegraphics[width=\imgwidth]{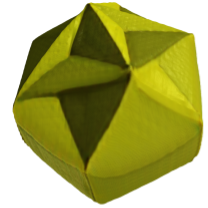} &
      \includegraphics[width=\imgwidth]{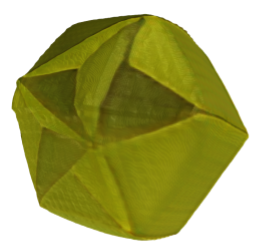} \\
    \hline
    \includegraphics[width=\imgwidth]{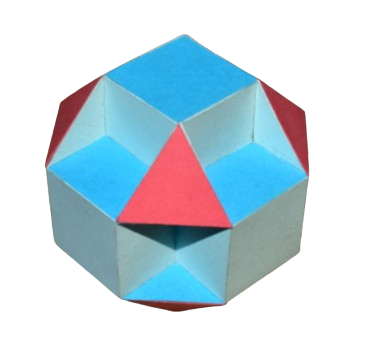} &
      \includegraphics[width=\imgwidth]{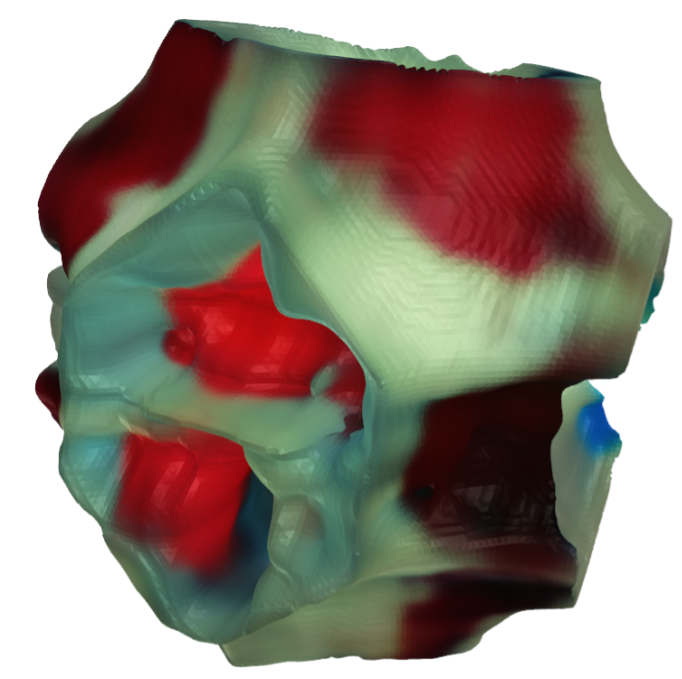} &
      \includegraphics[width=\imgwidth]{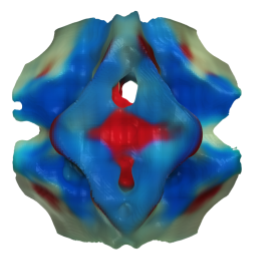} &
      \includegraphics[width=\imgwidth]{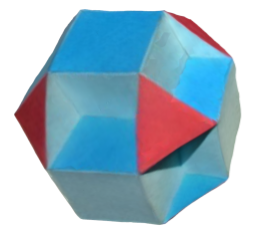} &
      \includegraphics[width=\imgwidth]{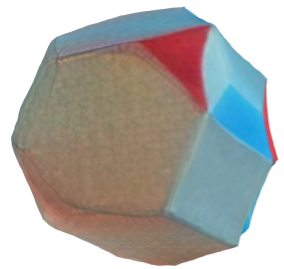} &
      \includegraphics[width=\imgwidth]{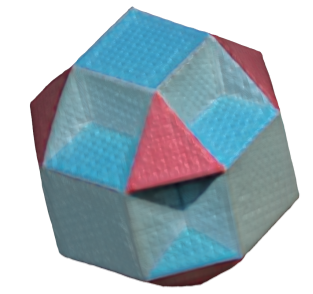} &
      \includegraphics[width=\imgwidth]{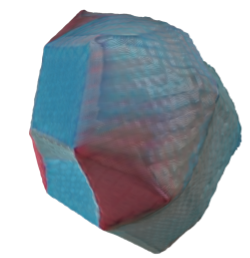} \\
    \hline
  \end{tabular}
\caption{Monocular 3D reconstruction results. Each method reconstructs a 3D shape from the input image, which we visualize by rendering the output from selected viewpoints. Rows depict pairs of synthetic and wild images of a cube (platonic solid), great dodecahedron (Kepler-Poinsot solid), and small cubicuboctahedron (uniform nonconvex solid).}
  \label{tab:m3dr-threeviews}
\end{table}

%% file: table_0shot.tex
\begin{figure}[!ht]
  \centering

  \begin{subfigure}[t]{0.48\textwidth}
    \centering
    \includegraphics[width=\textwidth,height=0.5\textheight,keepaspectratio,trim=0.3cm 5cm 1cm 3cm,clip]{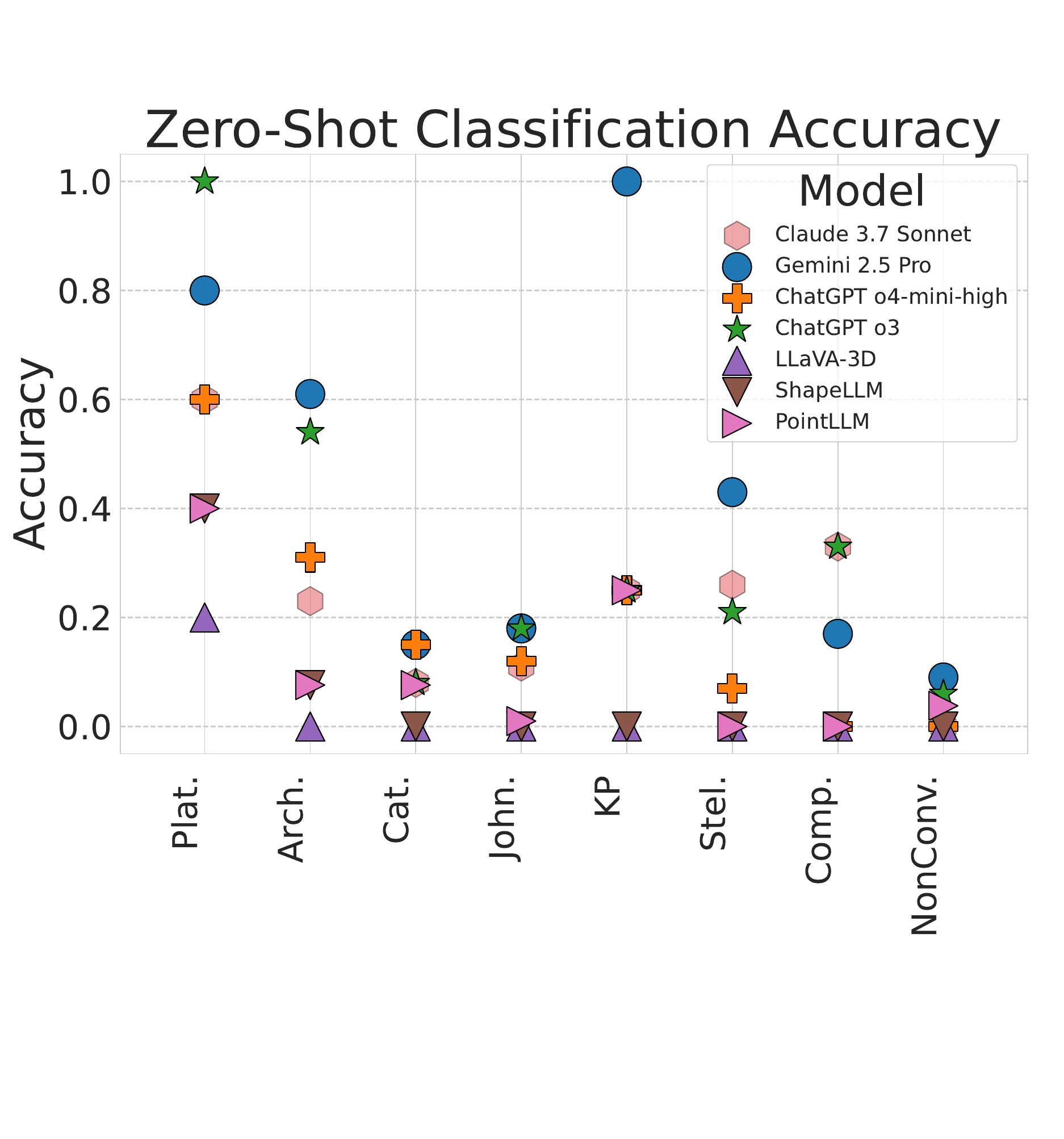}
    \caption{}
    \label{fig:0shot_half}
  \end{subfigure}%
  \hfill
  \begin{subfigure}[t]{0.48\textwidth}
    \centering
    \includegraphics[width=\textwidth,height=0.5\textheight,keepaspectratio]{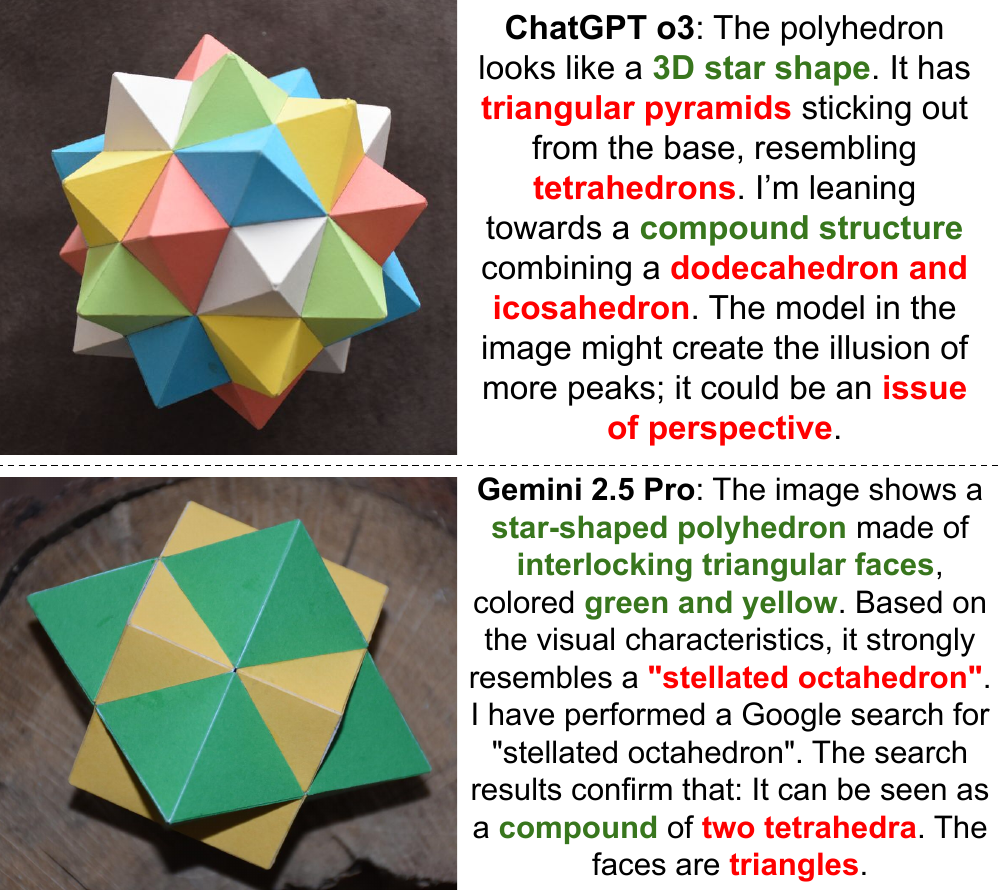}
    \caption{}
    \label{fig:0shotcls_reason}
  \end{subfigure}

  \caption{(a) Zero-shot classification accuracy of various frontier models across polyhedron categories using wild images. Results on synthetic images showed only marginal differences and are provided in the appendix. (b) Qualitative reasoning failures of frontier vision-language models. Correct text in {\textcolor{green!50!black}{\textbf{green}}}, incorrect in {\textcolor{red!70!black}{\textbf{red}}}.}
  \label{fig:combined_cls}
\end{figure}

%% file: 5-conclusion.tex
\section{Conclusion}

We introduced \dataset{}, a novel benchmark dataset designed to systematically assess the geometric reasoning capabilities of contemporary foundation vision and vision-language models using both synthetic and real-world polyhedral structures. Our extensive evaluations reveal notable discrepancies in current state-of-the-art models, highlighting a substantial disconnect between their impressive capabilities on standard benchmarks and their performance in tasks requiring explicit geometric reasoning. 

The results present a nuanced picture. On one hand, our 3D symmetry detection experiments show that pretrained vision encoders like DINOv2 can be surprisingly effective, as their image embeddings inherently capture fundamental 3D structural properties without explicit 3D supervision. This demonstrates the potential of foundation models to implicitly encode 3D structure and opens the door to utilizing them as lightweight, plug-and-play modules in downstream tasks.
On the other hand, this implicit understanding does not translate to robust, explicit reasoning in other domains. Our experiments underscore critical limitations in current models. 
Monocular 3D reconstruction approaches consistently failed to capture fundamental geometric properties from single images, even when trained on extensive, diverse datasets containing similar geometric primitives. \reb{Likely, these models learn priors for noisy, imperfect surfaces from their training data rather than mathematical exactness, suggesting that scale alone is insufficient without mathematically generated data to enforce geometric constraints.}
\reb{Similarly, mental rotation tests revealed an inability to capture subtle geometric distinctions. While the strongest model achieved parity with the average human baseline using non-linear probe, the vast majority of architectures faltered significantly. Crucially, our user study exposes a persistent gap in robust reasoning: a substantial majority of human participants outperformed even the best model, with top subjects demonstrating a decisive advantage on complex geometric instances.}
This was further confirmed by our zero-shot classification experiments, which revealed systematic errors in frontier models; they frequently misidentified fundamental properties such as face geometry, convexity, and compound structures, particularly within complex non-convex polyhedral classes.

\reb{We view \dataset{} as a ``geometric litmus test'' for spatial intelligence. Because geometric primitives are the building blocks of complex reasoning, failure here implies a fragile, texture-based approximation rather than true understanding. Real-world tasks rely on this same ``geometric arithmetic''; thus, models failing \dataset{} are unprepared for the ``geometric calculus'' of complex scenes. While establishing a precise correlation with arbitrary organic shapes requires further empirical study, \dataset{} isolates necessary foundational skills, paving the way for future rigorous evaluations of both rigid and fluid dynamics.}

\reb{Looking forward, \dataset{} offers a promising pathway for improving geometric intelligence. Its structured ontology enables the automatic generation of precise Chain-of-Thought (CoT) training data for step-by-step reasoning, though future training-focused extensions must scale to broader wild distributions to ensure robust generalization. Ultimately, \dataset{} serves as a critical diagnostic tool, providing the framework needed to drive the development of robust, geometry-aware foundation models capable of human-level spatial reasoning.}

\noindent \textbf{Ethics statement:} Authors used LLMs to help improve grammar and wording during the preparation of this manuscript.

\noindent\textbf{Acknowledgement.} Supported by the Intelligence Advanced Research Projects Activity (IARPA) via Department of
Interior/ Interior Business Center (DOI/IBC) contract number 140D0423C0076. The U.S.
Government is authorized to reproduce and distribute reprints for Governmental purposes
notwithstanding any copyright annotation thereon. Disclaimer: The views and conclusions
contained herein are those of the authors and should not be interpreted as necessarily
representing the official policies or endorsements, either expressed or implied, of IARPA,
DOI/IBC, or the U.S. Government.

We thank the students from Adam Mickiewicz High School in Kluczbork, Poland, for their valuable contribution in constructing the Johnson solids used in this work.

%% file: main.bbl
\begin{thebibliography}{64}
\providecommand{\natexlab}[1]{#1}
\providecommand{\url}[1]{\texttt{#1}}
\expandafter\ifx\csname urlstyle\endcsname\relax
  \providecommand{\doi}[1]{doi: #1}\else
  \providecommand{\doi}{doi: \begingroup \urlstyle{rm}\Url}\fi

\bibitem[Alain \& Bengio(2017)Alain and Bengio]{alain2017understanding}
Guillaume Alain and Yoshua Bengio.
\newblock Understanding intermediate layers using linear classifier probes, 2017.

\bibitem[Anthropic(2025)]{anthropic2025claude37}
Anthropic.
\newblock Claude 3.7 sonnet and claude code.
\newblock \url{https://www.anthropic.com/news/claude-3-7-sonnet}, Feb 2025.

\bibitem[Boss et~al.(2024)Boss, Huang, Vasishta, and Jampani]{boss2024sf3d}
Mark Boss, Zixuan Huang, Aaryaman Vasishta, and Varun Jampani.
\newblock Sf3d: Stable fast 3d mesh reconstruction with uv-unwrapping and illumination disentanglement.
\newblock \emph{arXiv preprint arXiv:2408.00653}, 2024.

\bibitem[Boss et~al.(2025)Boss, Huang, Vasishta, and Jampani]{boss2025sf3d}
Mark Boss, Zixuan Huang, Aaryaman Vasishta, and Varun Jampani.
\newblock Sf3d: Stable fast 3d mesh reconstruction with uv-unwrapping and illumination disentanglement.
\newblock In \emph{Proceedings of the Computer Vision and Pattern Recognition Conference}, pp.\  16240--16250, 2025.

\bibitem[Caron et~al.(2021)Caron, Touvron, Misra, J{\'e}gou, Mairal, Bojanowski, and Joulin]{caron2021emerging}
Mathilde Caron, Hugo Touvron, Ishan Misra, Herv{\'e} J{\'e}gou, Julien Mairal, Piotr Bojanowski, and Armand Joulin.
\newblock Emerging properties in self-supervised vision transformers.
\newblock In \emph{Proceedings of the IEEE/CVF international conference on computer vision}, pp.\  9650--9660, 2021.

\bibitem[Catalan(1865)]{catalan1865memoire}
Eugene Catalan.
\newblock M{\'e}moire sur la th{\'e}orie des poly{\`e}dres.
\newblock \emph{Journal de l'{\'E}cole Polytechnique}, 24, 1865.

\bibitem[Chen et~al.(2021)Chen, Hsieh, and Gong]{chen2021vision}
Xiangning Chen, Cho-Jui Hsieh, and Boqing Gong.
\newblock When vision transformers outperform resnets without pre-training or strong data augmentations.
\newblock \emph{arXiv preprint arXiv:2106.01548}, 2021.

\bibitem[Coxeter et~al.(1954)Coxeter, Longuet-Higgins, and Miller]{coxeter1954uniform}
Harold Scott~Macdonald Coxeter, Michael~Selwyn Longuet-Higgins, and Jeffery~CP Miller.
\newblock Uniform polyhedra.
\newblock \emph{Philosophical Transactions of the Royal Society of London. Series A, Mathematical and Physical Sciences}, 246\penalty0 (916):\penalty0 401--450, 1954.

\bibitem[Coxeter et~al.(2012)Coxeter, Flather, and Petrie]{coxeter2012fifty}
Harold Scott~Macdonald Coxeter, HT~Flather, and JF~Petrie.
\newblock \emph{The fifty-nine icosahedra}.
\newblock Springer Science \& Business Media, 2012.

\bibitem[DeepMind(2025)]{google2025gemini25pro}
Google DeepMind.
\newblock Gemini 2.5 pro.
\newblock \url{https://blog.google/technology/google-deepmind/gemini-model-thinking-updates-march-2025/}, Mar 2025.

\bibitem[Deitke et~al.(2023{\natexlab{a}})Deitke, Liu, Wallingford, Ngo, Michel, Kusupati, Fan, Laforte, Voleti, Gadre, et~al.]{deitke2023objaverse}
Matt Deitke, Ruoshi Liu, Matthew Wallingford, Huong Ngo, Oscar Michel, Aditya Kusupati, Alan Fan, Christian Laforte, Vikram Voleti, Samir~Yitzhak Gadre, et~al.
\newblock Objaverse-xl: A universe of 10m+ 3d objects.
\newblock \emph{Advances in Neural Information Processing Systems}, 36:\penalty0 35799--35813, 2023{\natexlab{a}}.

\bibitem[Deitke et~al.(2023{\natexlab{b}})Deitke, Schwenk, Salvador, Weihs, Michel, VanderBilt, Schmidt, Ehsani, Kembhavi, and Farhadi]{deitke2023objaversee}
Matt Deitke, Dustin Schwenk, Jordi Salvador, Luca Weihs, Oscar Michel, Eli VanderBilt, Ludwig Schmidt, Kiana Ehsani, Aniruddha Kembhavi, and Ali Farhadi.
\newblock Objaverse: A universe of annotated 3d objects.
\newblock In \emph{Proceedings of the IEEE/CVF conference on computer vision and pattern recognition}, pp.\  13142--13153, 2023{\natexlab{b}}.

\bibitem[Doveh et~al.(2023)Doveh, Arbelle, Harary, Herzig, Kim, Cascante-Bonilla, Alfassy, Panda, Giryes, Feris, et~al.]{doveh2023dense}
Sivan Doveh, Assaf Arbelle, Sivan Harary, Roei Herzig, Donghyun Kim, Paola Cascante-Bonilla, Amit Alfassy, Rameswar Panda, Raja Giryes, Rogerio Feris, et~al.
\newblock Dense and aligned captions (dac) promote compositional reasoning in vl models.
\newblock \emph{Advances in Neural Information Processing Systems}, 36:\penalty0 76137--76150, 2023.

\bibitem[Downs et~al.(2022)Downs, Francis, Koenig, Kinman, Hickman, Reymann, McHugh, and Vanhoucke]{downs2022google}
Laura Downs, Anthony Francis, Nate Koenig, Brandon Kinman, Ryan Hickman, Krista Reymann, Thomas~B McHugh, and Vincent Vanhoucke.
\newblock Google scanned objects: A high-quality dataset of 3d scanned household items.
\newblock In \emph{2022 International Conference on Robotics and Automation (ICRA)}, pp.\  2553--2560. IEEE, 2022.

\bibitem[El~Banani et~al.(2024)El~Banani, Raj, Maninis, Kar, Li, Rubinstein, Sun, Guibas, Johnson, and Jampani]{el2024probing}
Mohamed El~Banani, Amit Raj, Kevis-Kokitsi Maninis, Abhishek Kar, Yuanzhen Li, Michael Rubinstein, Deqing Sun, Leonidas Guibas, Justin Johnson, and Varun Jampani.
\newblock Probing the 3d awareness of visual foundation models.
\newblock In \emph{Proceedings of the IEEE/CVF Conference on Computer Vision and Pattern Recognition}, pp.\  21795--21806, 2024.

\bibitem[Euler(1758)]{euler1758elementa}
Leonhard Euler.
\newblock Elementa doctrinae solidorum.
\newblock \emph{Novi commentarii academiae scientiarum Petropolitanae}, pp.\  109--140, 1758.

\bibitem[Foret et~al.(2020)Foret, Kleiner, Mobahi, and Neyshabur]{foret2020sharpness}
Pierre Foret, Ariel Kleiner, Hossein Mobahi, and Behnam Neyshabur.
\newblock Sharpness-aware minimization for efficiently improving generalization.
\newblock \emph{arXiv preprint arXiv:2010.01412}, 2020.

\bibitem[Fu et~al.(2023)Fu, Tamir, Sundaram, Chai, Zhang, Dekel, and Isola]{fu2023dreamsim}
Stephanie Fu, Netanel Tamir, Shobhita Sundaram, Lucy Chai, Richard Zhang, Tali Dekel, and Phillip Isola.
\newblock Dreamsim: Learning new dimensions of human visual similarity using synthetic data.
\newblock \emph{arXiv preprint arXiv:2306.09344}, 2023.

\bibitem[Funk \& Liu(2017)Funk and Liu]{funk2017beyond}
Christopher Funk and Yanxi Liu.
\newblock Beyond planar symmetry: Modeling human perception of reflection and rotation symmetries in the wild.
\newblock In \emph{Proceedings of the IEEE international conference on computer vision}, pp.\  793--803, 2017.

\bibitem[Gao et~al.(2020)Gao, Zhang, Meng, Ren, Lai, and Kobbelt]{gao2020prs}
Lin Gao, Ling-Xiao Zhang, Hsien-Yu Meng, Yi-Hui Ren, Yu-Kun Lai, and Leif Kobbelt.
\newblock Prs-net: Planar reflective symmetry detection net for 3d models.
\newblock \emph{IEEE transactions on visualization and computer graphics}, 27\penalty0 (6):\penalty0 3007--3018, 2020.

\bibitem[Guo et~al.(2023)Guo, Zhang, Zhu, Tang, Ma, Han, Chen, Gao, Li, Li, et~al.]{guo2023point}
Ziyu Guo, Renrui Zhang, Xiangyang Zhu, Yiwen Tang, Xianzheng Ma, Jiaming Han, Kexin Chen, Peng Gao, Xianzhi Li, Hongsheng Li, et~al.
\newblock Point-bind \& point-llm: Aligning point cloud with multi-modality for 3d understanding, generation, and instruction following.
\newblock \emph{arXiv preprint arXiv:2309.00615}, 2023.

\bibitem[He et~al.(2022)He, Chen, Xie, Li, Doll{\'a}r, and Girshick]{he2022masked}
Kaiming He, Xinlei Chen, Saining Xie, Yanghao Li, Piotr Doll{\'a}r, and Ross Girshick.
\newblock Masked autoencoders are scalable vision learners.
\newblock In \emph{Proceedings of the IEEE/CVF conference on computer vision and pattern recognition}, pp.\  16000--16009, 2022.

\bibitem[He \& Wang(2023)He and Wang]{openlrm}
Zexin He and Tengfei Wang.
\newblock Openlrm: Open-source large reconstruction models.
\newblock \url{https://github.com/3DTopia/OpenLRM}, 2023.

\bibitem[Hong et~al.(2023)Hong, Zhang, Gu, Bi, Zhou, Liu, Liu, Sunkavalli, Bui, and Tan]{hong2023lrm}
Yicong Hong, Kai Zhang, Jiuxiang Gu, Sai Bi, Yang Zhou, Difan Liu, Feng Liu, Kalyan Sunkavalli, Trung Bui, and Hao Tan.
\newblock Lrm: Large reconstruction model for single image to 3d.
\newblock \emph{arXiv preprint arXiv:2311.04400}, 2023.

\bibitem[Hong et~al.(2024)Hong, Zhang, Gu, Bi, Zhou, Liu, Liu, Sunkavalli, Bui, and Tan]{honglrm}
Yicong Hong, Kai Zhang, Jiuxiang Gu, Sai Bi, Yang Zhou, Difan Liu, Feng Liu, Kalyan Sunkavalli, Trung Bui, and Hao Tan.
\newblock Lrm: Large reconstruction model for single image to 3d.
\newblock In \emph{The Twelfth International Conference on Learning Representations}, 2024.

\bibitem[Jampani et~al.(2023)Jampani, Maninis, Engelhardt, Karpur, Truong, Sargent, Popov, Araujo, Martin~Brualla, Patel, et~al.]{jampani2023navi}
Varun Jampani, Kevis-Kokitsi Maninis, Andreas Engelhardt, Arjun Karpur, Karen Truong, Kyle Sargent, Stefan Popov, Andr{\'e} Araujo, Ricardo Martin~Brualla, Kaushal Patel, et~al.
\newblock Navi: Category-agnostic image collections with high-quality 3d shape and pose annotations.
\newblock \emph{Advances in Neural Information Processing Systems}, 36:\penalty0 76061--76084, 2023.

\bibitem[Johnson(1966)]{Johnson1966}
N.~W. Johnson.
\newblock Convex polyhedra with regular faces.
\newblock \emph{Canadian Journal of Mathematics}, 18:\penalty0 169--200, 1966.

\bibitem[Jun \& Nichol(2023)Jun and Nichol]{jun2023shap}
Heewoo Jun and Alex Nichol.
\newblock Shap-e: Generating conditional 3d implicit functions.
\newblock \emph{arXiv preprint arXiv:2305.02463}, 2023.

\bibitem[Kepler(1619)]{kepler1969harmonices}
Johannes Kepler.
\newblock \emph{Harmonices mundi libri V}.
\newblock 1619.

\bibitem[Klein(2003)]{klein2003lectures}
Felix Klein.
\newblock \emph{Lectures on the Icosahedron and the Solution of Equations of the Fifth Degree}.
\newblock Courier Corporation, 2003.

\bibitem[Leroy et~al.(2024)Leroy, Cabon, and Revaud]{leroy2024grounding}
Vincent Leroy, Yohann Cabon, and J{\'e}r{\^o}me Revaud.
\newblock Grounding image matching in 3d with mast3r.
\newblock In \emph{European Conference on Computer Vision}, pp.\  71--91. Springer, 2024.

\bibitem[Lewis et~al.(2022)Lewis, Nayak, Yu, Yu, Merullo, Bach, and Pavlick]{lewis2022does}
Martha Lewis, Nihal~V Nayak, Peilin Yu, Qinan Yu, Jack Merullo, Stephen~H Bach, and Ellie Pavlick.
\newblock Does clip bind concepts? probing compositionality in large image models.
\newblock \emph{arXiv preprint arXiv:2212.10537}, 2022.

\bibitem[Liu et~al.(2022)Liu, Mao, Wu, Feichtenhofer, Darrell, and Xie]{liu2022convnet}
Zhuang Liu, Hanzi Mao, Chao-Yuan Wu, Christoph Feichtenhofer, Trevor Darrell, and Saining Xie.
\newblock A convnet for the 2020s.
\newblock In \emph{Proceedings of the IEEE/CVF conference on computer vision and pattern recognition}, pp.\  11976--11986, 2022.

\bibitem[Man et~al.(2024)Man, Zheng, Bao, Hebert, Gui, and Wang]{man2024lexicon3d}
Yunze Man, Shuhong Zheng, Zhipeng Bao, Martial Hebert, Liangyan Gui, and Yu-Xiong Wang.
\newblock Lexicon3d: Probing visual foundation models for complex 3d scene understanding.
\newblock \emph{Advances in Neural Information Processing Systems}, 37:\penalty0 76819--76847, 2024.

\bibitem[Nguyen et~al.(2022)Nguyen, Ilharco, Wortsman, Oh, and Schmidt]{nguyen2022quality}
Thao Nguyen, Gabriel Ilharco, Mitchell Wortsman, Sewoong Oh, and Ludwig Schmidt.
\newblock Quality not quantity: On the interaction between dataset design and robustness of clip.
\newblock \emph{Advances in Neural Information Processing Systems}, 35:\penalty0 21455--21469, 2022.

\bibitem[Nimier-David et~al.(2019)Nimier-David, Vicini, Zeltner, and Jakob]{nimier2019mitsuba}
Merlin Nimier-David, Delio Vicini, Tizian Zeltner, and Wenzel Jakob.
\newblock Mitsuba 2: A retargetable forward and inverse renderer.
\newblock \emph{ACM Transactions on Graphics (ToG)}, 38\penalty0 (6):\penalty0 1--17, 2019.

\bibitem[OpenAI(2025)]{openai2025o3}
OpenAI.
\newblock Introducing openai o3 and o4-mini.
\newblock \url{https://openai.com/index/introducing-o3-and-o4-mini/}, Apr 2025.

\bibitem[Oquab et~al.(2023)Oquab, Darcet, Moutakanni, Vo, Szafraniec, Khalidov, Fernandez, Haziza, Massa, El-Nouby, et~al.]{oquab2023dinov2}
Maxime Oquab, Timoth{\'e}e Darcet, Th{\'e}o Moutakanni, Huy Vo, Marc Szafraniec, Vasil Khalidov, Pierre Fernandez, Daniel Haziza, Francisco Massa, Alaaeldin El-Nouby, et~al.
\newblock Dinov2: Learning robust visual features without supervision.
\newblock \emph{arXiv preprint arXiv:2304.07193}, 2023.

\bibitem[Qi et~al.(2024)Qi, Dong, Zhang, Geng, Han, Ge, Yi, and Ma]{qi2024shapellm}
Zekun Qi, Runpei Dong, Shaochen Zhang, Haoran Geng, Chunrui Han, Zheng Ge, Li~Yi, and Kaisheng Ma.
\newblock Shapellm: Universal 3d object understanding for embodied interaction.
\newblock In \emph{European Conference on Computer Vision}, pp.\  214--238. Springer, 2024.

\bibitem[Radford et~al.(2021)Radford, Kim, Hallacy, Ramesh, Goh, Agarwal, Sastry, Askell, Mishkin, Clark, et~al.]{radford2021learning}
Alec Radford, Jong~Wook Kim, Chris Hallacy, Aditya Ramesh, Gabriel Goh, Sandhini Agarwal, Girish Sastry, Amanda Askell, Pamela Mishkin, Jack Clark, et~al.
\newblock Learning transferable visual models from natural language supervision.
\newblock In \emph{International conference on machine learning}, pp.\  8748--8763. PMLR, 2021.

\bibitem[Ramakrishnan et~al.(2024)Ramakrishnan, Wijmans, Kraehenbuehl, and Koltun]{ramakrishnan2024does}
Santhosh~Kumar Ramakrishnan, Erik Wijmans, Philipp Kraehenbuehl, and Vladlen Koltun.
\newblock Does spatial cognition emerge in frontier models?
\newblock \emph{arXiv preprint arXiv:2410.06468}, 2024.

\bibitem[Schlarmann \& Hein(2023)Schlarmann and Hein]{schlarmann2023adversarial}
Christian Schlarmann and Matthias Hein.
\newblock On the adversarial robustness of multi-modal foundation models.
\newblock In \emph{Proceedings of the IEEE/CVF International Conference on Computer Vision}, pp.\  3677--3685, 2023.

\bibitem[Shepard \& Metzler(1971)Shepard and Metzler]{shepard1971mental}
Roger~N Shepard and Jacqueline Metzler.
\newblock Mental rotation of three-dimensional objects.
\newblock \emph{Science}, 171\penalty0 (3972):\penalty0 701--703, 1971.

\bibitem[Stewart(1980)]{stewart1980adventures}
Bonnie Stewart.
\newblock Adventures among the toroids.
\newblock \emph{Structural Topology, 1980, n{\'u}m. 5}, 1980.

\bibitem[Tang et~al.(2024)Tang, Qu, Wang, Zhuang, Wu, Ma, Wang, Zheng, Zhao, and Zhao]{tang2024sparkle}
Yihong Tang, Ao~Qu, Zhaokai Wang, Dingyi Zhuang, Zhaofeng Wu, Wei Ma, Shenhao Wang, Yunhan Zheng, Zhan Zhao, and Jinhua Zhao.
\newblock Sparkle: Mastering basic spatial capabilities in vision language models elicits generalization to composite spatial reasoning.
\newblock \emph{arXiv preprint arXiv:2410.16162}, 2024.

\bibitem[Tochilkin et~al.(2024)Tochilkin, Pankratz, Liu, Huang, Letts, Li, Liang, Laforte, Jampani, and Cao]{tochilkin2024triposr}
Dmitry Tochilkin, David Pankratz, Zexiang Liu, Zixuan Huang, Adam Letts, Yangguang Li, Ding Liang, Christian Laforte, Varun Jampani, and Yan-Pei Cao.
\newblock Triposr: Fast 3d object reconstruction from a single image.
\newblock \emph{arXiv preprint arXiv:2403.02151}, 2024.

\bibitem[Tong et~al.(2024)Tong, Brown, Wu, Woo, IYER, Akula, Yang, Yang, Middepogu, Wang, et~al.]{tong2024cambrian}
Peter Tong, Ellis Brown, Penghao Wu, Sanghyun Woo, Adithya Jairam~Vedagiri IYER, Sai~Charitha Akula, Shusheng Yang, Jihan Yang, Manoj Middepogu, Ziteng Wang, et~al.
\newblock Cambrian-1: A fully open, vision-centric exploration of multimodal llms.
\newblock \emph{Advances in Neural Information Processing Systems}, 37:\penalty0 87310--87356, 2024.

\bibitem[Touvron et~al.(2022)Touvron, Cord, and J{\'e}gou]{touvron2022deit}
Hugo Touvron, Matthieu Cord, and Herv{\'e} J{\'e}gou.
\newblock Deit iii: Revenge of the vit.
\newblock In \emph{European conference on computer vision}, pp.\  516--533. Springer, 2022.

\bibitem[Touvron et~al.(2023)Touvron, Lavril, Izacard, Martinet, Lachaux, Lacroix, Rozi{\`e}re, Goyal, Hambro, Azhar, et~al.]{touvron2023llama}
Hugo Touvron, Thibaut Lavril, Gautier Izacard, Xavier Martinet, Marie-Anne Lachaux, Timoth{\'e}e Lacroix, Baptiste Rozi{\`e}re, Naman Goyal, Eric Hambro, Faisal Azhar, et~al.
\newblock Llama: Open and efficient foundation language models.
\newblock \emph{arXiv preprint arXiv:2302.13971}, 2023.

\bibitem[Tsogkas \& Kokkinos(2012)Tsogkas and Kokkinos]{tsogkas2012learning}
Stavros Tsogkas and Iasonas Kokkinos.
\newblock Learning-based symmetry detection in natural images.
\newblock In \emph{Computer Vision--ECCV 2012: 12th European Conference on Computer Vision, Florence, Italy, October 7-13, 2012, Proceedings, Part VII 12}, pp.\  41--54. Springer, 2012.

\bibitem[Vandenberg \& Kuse(1978)Vandenberg and Kuse]{vandenberg1978mental}
Steven~G Vandenberg and Allan~R Kuse.
\newblock Mental rotations, a group test of three-dimensional spatial visualization.
\newblock \emph{Perceptual and motor skills}, 47\penalty0 (2):\penalty0 599--604, 1978.

\bibitem[Verheyen(1989)]{verheyen1989complete}
Hugo~F Verheyen.
\newblock The complete set of jitterbug transformers and the analysis of their motion.
\newblock In \emph{Symmetry 2}, pp.\  203--250. Elsevier, 1989.

\bibitem[Wagemans(1997)]{wagemans1997characteristics}
Johan Wagemans.
\newblock Characteristics and models of human symmetry detection.
\newblock \emph{Trends in cognitive sciences}, 1\penalty0 (9):\penalty0 346--352, 1997.

\bibitem[Wang et~al.(2025)Wang, Chen, Karaev, Vedaldi, Rupprecht, and Novotny]{wang2025vggt}
Jianyuan Wang, Minghao Chen, Nikita Karaev, Andrea Vedaldi, Christian Rupprecht, and David Novotny.
\newblock Vggt: Visual geometry grounded transformer.
\newblock In \emph{Proceedings of the Computer Vision and Pattern Recognition Conference}, pp.\  5294--5306, 2025.

\bibitem[Wang et~al.(2024)Wang, Leroy, Cabon, Chidlovskii, and Revaud]{wang2024dust3r}
Shuzhe Wang, Vincent Leroy, Yohann Cabon, Boris Chidlovskii, and Jerome Revaud.
\newblock Dust3r: Geometric 3d vision made easy.
\newblock In \emph{Proceedings of the IEEE/CVF Conference on Computer Vision and Pattern Recognition}, pp.\  20697--20709, 2024.

\bibitem[Wenninger(1971)]{Wenninger1971}
M.~J. Wenninger.
\newblock \emph{Polyhedron Models}.
\newblock Cambridge University Press, 1971.

\bibitem[Wu et~al.(2023)Wu, Zhang, Fu, Wang, Ren, Pan, Wu, Yang, Wang, Qian, et~al.]{wu2023omniobject3d}
Tong Wu, Jiarui Zhang, Xiao Fu, Yuxin Wang, Jiawei Ren, Liang Pan, Wayne Wu, Lei Yang, Jiaqi Wang, Chen Qian, et~al.
\newblock Omniobject3d: Large-vocabulary 3d object dataset for realistic perception, reconstruction and generation.
\newblock In \emph{Proceedings of the IEEE/CVF Conference on Computer Vision and Pattern Recognition}, pp.\  803--814, 2023.

\bibitem[Yamada et~al.(2023)Yamada, Bao, Lampinen, Kasai, and Yildirim]{yamada2023evaluating}
Yutaro Yamada, Yihan Bao, Andrew~K Lampinen, Jungo Kasai, and Ilker Yildirim.
\newblock Evaluating spatial understanding of large language models.
\newblock \emph{arXiv preprint arXiv:2310.14540}, 2023.

\bibitem[Yang et~al.(2023)Yang, Zhang, Li, Zou, Li, and Gao]{yang2023set}
Jianwei Yang, Hao Zhang, Feng Li, Xueyan Zou, Chunyuan Li, and Jianfeng Gao.
\newblock Set-of-mark prompting unleashes extraordinary visual grounding in gpt-4v.
\newblock \emph{arXiv preprint arXiv:2310.11441}, 2023.

\bibitem[Yu et~al.(2023)Yu, Xu, Zhang, Liu, Ye, Wu, Yan, Zhu, Xiong, Liang, et~al.]{yu2023mvimgnet}
Xianggang Yu, Mutian Xu, Yidan Zhang, Haolin Liu, Chongjie Ye, Yushuang Wu, Zizheng Yan, Chenming Zhu, Zhangyang Xiong, Tianyou Liang, et~al.
\newblock Mvimgnet: A large-scale dataset of multi-view images.
\newblock In \emph{Proceedings of the IEEE/CVF conference on computer vision and pattern recognition}, pp.\  9150--9161, 2023.

\bibitem[Zalgaller(1969)]{zalgaller1969convex}
Viktor~A Zalgaller.
\newblock Convex polyhedra with regular faces.
\newblock \emph{(No Title)}, 1969.

\bibitem[Zhai et~al.(2023)Zhai, Mustafa, Kolesnikov, and Beyer]{zhai2023sigmoid}
Xiaohua Zhai, Basil Mustafa, Alexander Kolesnikov, and Lucas Beyer.
\newblock Sigmoid loss for language image pre-training.
\newblock In \emph{Proceedings of the IEEE/CVF International Conference on Computer Vision}, pp.\  11975--11986, 2023.

\bibitem[Zhu et~al.(2024{\natexlab{a}})Zhu, Wang, Zhang, Pang, and Liu]{zhu2024llava}
Chenming Zhu, Tai Wang, Wenwei Zhang, Jiangmiao Pang, and Xihui Liu.
\newblock Llava-3d: A simple yet effective pathway to empowering lmms with 3d-awareness.
\newblock \emph{arXiv preprint arXiv:2409.18125}, 2024{\natexlab{a}}.

\bibitem[Zhu et~al.(2024{\natexlab{b}})Zhu, Cai, Deng, Ooi, and Wu]{zhu2024llms}
Jiaqi Zhu, Shaofeng Cai, Fang Deng, Beng~Chin Ooi, and Junran Wu.
\newblock Do llms understand visual anomalies? uncovering llm's capabilities in zero-shot anomaly detection.
\newblock In \emph{Proceedings of the 32nd ACM International Conference on Multimedia}, pp.\  48--57, 2024{\natexlab{b}}.

\end{thebibliography}
